\documentclass{article}
\usepackage{spconf,amsmath,graphicx}
\usepackage{algorithm}
\usepackage{algorithmic}
\usepackage{booktabs}
\usepackage{cite}
\usepackage{hyperref}
\usepackage{enumitem}
\usepackage{amssymb}
\usepackage{multirow}
\usepackage{makecell}
\usepackage{longtable}
\usepackage[table ]{ xcolor}
\usepackage{arydshln}
\usepackage{booktabs}
\usepackage{subcaption}
\usepackage{wrapfig}

\definecolor{mygray}{gray}{.9}
\setenumerate[1]{itemsep=0pt,partopsep=0pt,parsep=\parskip,topsep=0pt}
\setitemize[1]{itemsep=0pt,partopsep=0pt,parsep=\parskip,topsep=0pt}

\hypersetup{hidelinks,
	colorlinks=true,
	allcolors=black,
	pdfstartview=Fit,
	breaklinks=true}

\title{Diffusion-based Pose Refinement and Multi-Hypothesis Generation for 3D Human Pose Estimation}
\name{Hongbo Kang\textsuperscript{1},  Yong Wang\textsuperscript{*1}, Mengyuan Liu\textsuperscript{2}, Doudou Wu\textsuperscript{1}, Peng Liu\textsuperscript{1}, Xinlin Yuan\textsuperscript{1}, Wenming Yang\textsuperscript{3} \thanks{* Corresponding author. This work was partly supported by the National Natural Science Foundation of China(No.62171251) and the Special Foundations for the Development of Strategic Emerging Industries of Shenzhen(Nos.JCYJ20200109143035495, CJGJZD20210408092804011 \& JSGG20211108092812020).}}
\address{Chongqing University of Technology\textsuperscript{1} , Peking University\textsuperscript{2} , Tsinghua University\textsuperscript{3}\\
\texttt{\normalsize \{hbkang,doudouwu,52222313123,52212313128\}@stu.cqut.edu.cn, ywang@cqut.edu.cn,}\\
\texttt{\normalsize  liumengyuan@pku.edu.cn, yang.wenming@sz.tsinghua.edu.cn}}

\begin{document}
\ninept
\maketitle
\begin{abstract}

Previous probabilistic models for 3D Human Pose Estimation (3DHPE) aimed to enhance pose accuracy by generating multiple hypotheses. However, most of the hypotheses generated deviate substantially from the true pose. Compared to deterministic models, the excessive uncertainty in probabilistic models leads to weaker performance in single-hypothesis prediction. To address these two challenges, we propose a diffusion-based refinement framework called DRPose, which refines the output of deterministic models by reverse diffusion and achieves more suitable multi-hypothesis prediction for the current pose benchmark by multi-step refinement with multiple noises. To this end, we propose a Scalable Graph Convolution Transformer (SGCT) and a Pose Refinement Module (PRM) for denoising and refining. Extensive experiments on Human3.6M and MPI-INF-3DHP datasets demonstrate that our method achieves state-of-the-art performance on both single and multi-hypothesis 3DHPE. Code is available at \href{https://github.com/KHB1698/DRPose}{\textcolor{red}{https://github.com/KHB1698/DRPose}}.


%
\end{abstract}
\begin{keywords}
3D Human Pose Estimation, Diffusion Model, Pose Refinement, Multi-Hypothesis Generation
\end{keywords}


%
\section{Introduction}
\label{sec:intro}

In recent years, 3D Human Pose Estimation (3DHPE) has garnered widespread attention due to its significant applications in fields such as action recognition\cite{liu2018recognizing}, virtual reality\cite{mehta2017vnect}, and human-computer interaction\cite{preece1994human}. The primary objective of 3DHPE is to accurately predict the 3D poses of the human body from images or videos. Substantial progress has been achieved in 3D pose estimation from 2D pose inputs, facilitated by powerful 2D pose estimators\cite{newell2016stacked,chen2018cascaded,sun2019deep}.



A considerable body of work has focused on deterministic models\cite{pavllo20193d,zou2021modulated,zhang2022mixste,zhao2023poseformerv2,cai2023htnet,kang2023double} aimed at inferring the most likely pose directly from given input data. However, these models often grapple with the inherent ambiguity present in the data, leading to suboptimal results, particularly in complex and challenging scenarios. To address this issue, probabilistic models\cite{wehrbein2021probabilistic,oikarinen2021graphmdn,li2022mhformer,ci2023gfpose} have been introduced, aiming to capture the uncertainty in pose estimation by generating multiple pose hypotheses. By encompassing a broader range of potential solutions, this approach holds promise for more accurate pose predictions.

\begin{figure}[thb]
\centering
\includegraphics[width=\linewidth]{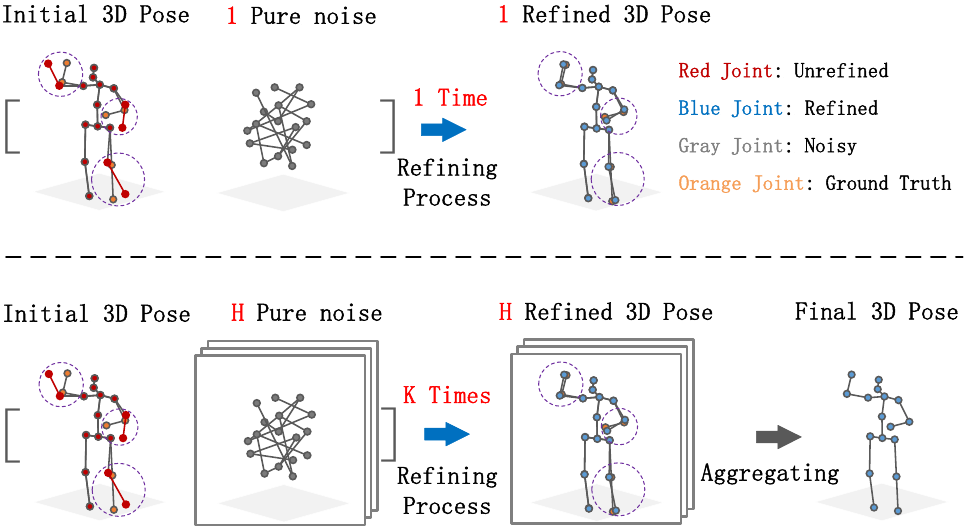}
\caption{Overview of the DRPose framework in the inference stage for pose refinement.{\bf{Top:}} Single-hypothesis inference. The initial 3D pose is combined with a pure noise and refined once to obtain the refined 3D pose. {\bf{Bottom:}} Multi-hypothesis inference. The initial 3D pose is combined with multiple pure noise and refined multiple times to obtain multiple refined 3D poses. In real-world applications, the final 3D pose is obtained through the aggregation from multi-hypothesis.}
\label{fig:inference}
\end{figure}

\begin{figure*}[thb]
\centering
\includegraphics[width=\textwidth]{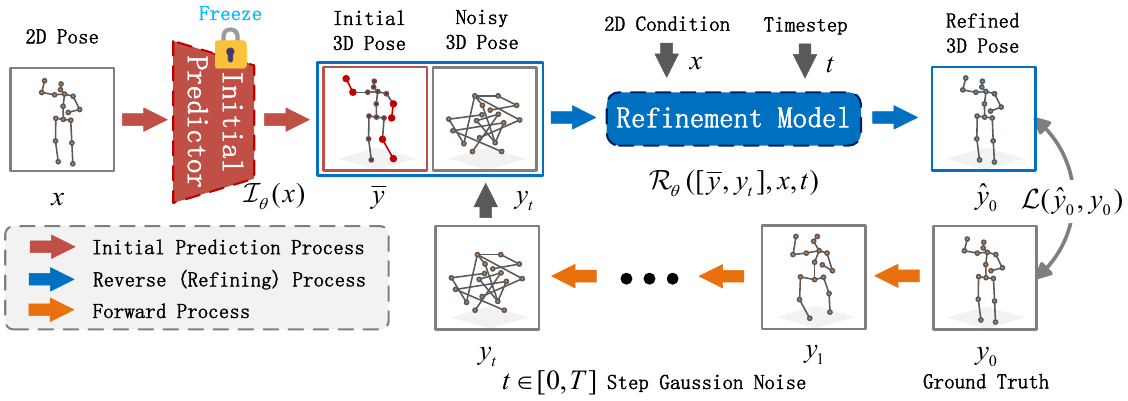}
\caption{Overview of the DRPose framework in the training stage. Through the forward process, the ground truth is diffused to obtain the noisy 3D pose, and it is combined with the initial 3D pose obtained by the initial predictor as the input of the reverse process. Then, using 2D pose and timestep as conditions to denoise and refine the input. The refined 3D pose is obtained at last.}
\label{fig:diffusion}
\end{figure*}


Although probabilistic models offer advantages in handling uncertainty, excessive uncertainty makes them face two main problems: ($\romannumeral1$) most generated hypotheses often exhibit significant deviations from the ground truth pose, resulting in averaged outcomes considerably inferior to the optimal solution; ($\romannumeral2$) under the context of single-pose hypothesis prediction, probabilistic models demonstrate weaker performance compared to deterministic models, impeding their widespread adoption. We contend that the key to addressing these challenges lies in aligning the average distribution of all hypotheses closer to the true values, enabling multi-hypothesis average outcomes to achieve comparable or even superior performance to deterministic models.

Consequently, we propose a diffusion-based\cite{ho2020denoising,song2020denoising} refinement framework, called DRPose, aimed at refining 3D poses generated by deterministic models\cite{cai2023htnet,kang2023double} to bring the average predictions of probabilistic models closer to the real distribution. As depicted in Fig. \ref{fig:inference}, we combine initial 3D poses (deterministic) predicted by the deterministic model with pure noise (probabilistic) to capture their underlying 3D features, achieving denoising and refinement effects. By introducing different noises, we obtain predictive results for multiple hypotheses. Importantly, each hypothesis's distribution still adheres to the characteristics of single-hypothesis predictions. This ensures that the multi-hypothesis average distribution obtained through DRPose is closer to the true values, and the iterative nature of the diffusion model further enhances differences between distinct hypotheses.

The core success of DRPose lies in two key components: a Scalable Graph Convolution Transformer (SGCT) and a Pose Refinement Module (PRM). SGCT is primarily used for denoising and learning the distribution of initial 3D poses and their latent features. Ultimately, through PRM, a balance is struck between certain and uncertain poses, yielding a refined final pose. The effective integration of these two models results in more robust and accurate 3D pose estimation.

Contributions in this paper can be summarized as follows: 


\begin{itemize}
	\item We propose a DRPose framework for refining 3D poses and achieving more accurate multi-hypothesis extensions.
	\item We design two innovative components: SGCT for denoising and learning the distribution of initial 3D poses and latent features, and PRM for balancing certain and uncertain poses.
	\item Extensive experiments validate the efficacy of DRPose. Our approach achieves state-of-the-art performance in both single-hypothesis and multi-hypothesis 3DHPE scenarios on the Human3.6M and MPI-INF-3DHP datasets.
  \end{itemize}

\section{Method}
\label{sec:method}


For training, as shown in Fig. \ref{fig:diffusion}, given the input 2D pose $x\in\mathbb{R}^{N\times2}$, where $N$ is the number of joints, the initial predictor is used to obtain the initial 3D pose $\bar{y}\in\mathbb{R}^{N\times3}$. The model performs $t$ steps of forward diffusion to obtain the noisy 3D pose $y_t\in\mathbb{R}^{N\times3}$ from the ground truth $y_0\in\mathbb{R}^{N\times3}$. Then, the deterministic $\bar{y}$ is combined with the probabilistic $y_t$ as input, and a refinement model (including SGCT and PRM) is used to obtain a more accurate 3D representation $\hat{y}_0\in\mathbb{R}^{N\times3}$. For inference, as shown in Fig. \ref{fig:inference}, the initial pose is combined with 1 noise and refined once to implement single-hypothesis prediction. In addition, the initial pose is combined with multiple noises and refined multiple times to implement multi-hypothesis prediction, and a single accurate 3D pose is generated for practical use by the aggregation method\cite{shan2023diffusion}.

\subsection{Diffusion-based Pose Refinement}
\label{ssec:d3dr}



Our DRPose is based on a diffusion model. For the forward process, the ground truth 3D pose is gradually disturbed by noise. For the reverse process, the noise is transformed back to the target distribution. Given a training sample $y_0 \sim q(y_0)$, the noisy versions $\{y_t\}_{t=1}^T$ are obtained according to the following Markov process:
\begin{equation}
\begin{split}
\label{eq:diffusion}
q(y_t|y_{t-1}) &:= \mathcal{N}(y_t;\sqrt[]{1-\beta_t}y_{t-1},\beta_tI)
\end{split}
\end{equation}
where $t=1,2,...,T$ and $\beta_t$ is the cosine noise variance schedule. The marginal distribution of $y_t$ is given by:
\begin{gather}
\label{eq:diffusion1}
q(y_t|y_0) := \mathcal{N}(y_t;\sqrt[]{\bar{\alpha }_t}y_0,(1-\bar{\alpha}_t)I)
\end{gather}
where $\alpha_t:=1-\beta_t$ and $\bar{\alpha}_t:=\prod_{s=1}^t \alpha_s$. For the reverse process, the distribution $q(y_{t-1}|y_t)$ is estimated by a neural network $\mathcal{R}_{\theta }$, which can be expressed as:
\begin{gather}
\label{eq:recovery}
p(y_{t-1}|y_t) := \mathcal{N}(y_{t-1};\mu_\theta(y_t,t),\Sigma_\theta(y_t,t))
\end{gather}
Although it is feasible to estimate $y_{t-1}$ directly, our goal is to obtain a more refined 3D pose, so reconstructing $y_0$ is more conducive to improving performance. In our proposed DRPose, we need to pre-train an initial predictor $\mathcal{I}_{\theta }$ to obtain our initial 3D pose $\bar{y}$, as follows:
\begin{gather}
\label{eq:initial}
\bar{y} = \mathcal{I}_{\theta } (x)
\end{gather}
Then, $\bar{y}$ is combined with the noisy version $y_t$ obtained by the forward process as an input to the network. In addition, we also use 2D pose $x$ and step $t$ as conditions to scale and shift our refinement model $\mathcal{R}_{\theta }$. Through training this network to predict the refined 3D pose $\hat{y}_0$, as follows:
\begin{gather}
\label{eq:refine}
\hat{y}_0 = \mathcal{R}_{\theta }([\bar{y},y_t],x,t)
\end{gather}
where $[\bar{y},y_t]$ denotes the concatenation of $\bar{y}$ and $y_t$. In practice, for Eq. (\ref{eq:recovery}), $\Sigma_\theta (y_t, t)$ is set to $\sigma^2_t = \frac{1-\bar{\alpha}_{t-1}}{1-\bar{\alpha}_t}\beta_t$, and $\mu_\theta (y_t, t)$ can be expressed as:
\begin{gather}
\label{eq:recovery1}
\mu_\theta(y_t,t)=\frac{\sqrt[]{\bar{\alpha}_t}(1-\bar{\alpha}_{t-1})}{1-\bar{\alpha}_t}y_t+\frac{\sqrt[]{\bar{\alpha}_{t-1}}\beta_t}{1-\bar{\alpha}_t}\hat{y}_0
\end{gather}
where $\hat{y}_0$ is the refined 3D pose by our refinement model as in Eq. (\ref{eq:refine}), which consists of SGCT and PRM. Then our framework is trained by minimizing the following loss function:
\begin{gather}
\label{eq:loss_w}
\mathcal{L}(\hat{y}_0,y_0) = \frac{1}{N} \sum_{i=1}^N(\lambda_i \Vert \hat{y}_{0,i} - y_{0,i} \Vert_2)
\end{gather}
where $y_{0,i}$ denotes the ground truth 3D joint position for joint $i$, while $\hat{y}_{0,i}$ represents its estimated counterpart. Whereas $\lambda_i$ represents the weighting factor for joint $i$.




\begin{figure}[t]
\centering
\includegraphics[width=.6\linewidth]{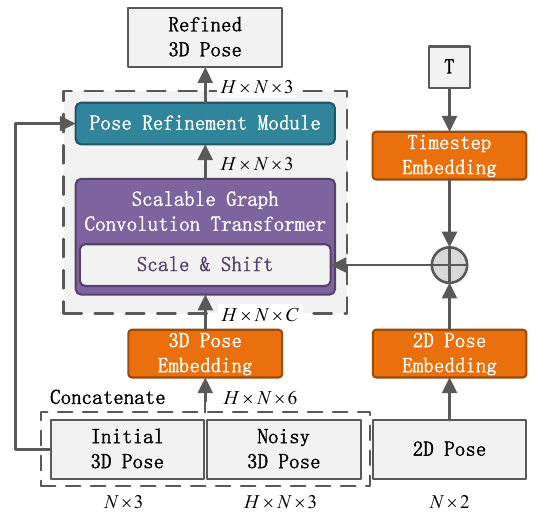}
\caption{Overview of the Refinement Model, which consists of the Scalable Graph Convolution Transformer (SGCT) and Pose Refinement Module (PRM).}
\label{fig:sgraformer}
\end{figure}

\subsection{Architecture}
\label{ssec:arch}

For the implementation of our framework, we design a refinement model $\mathcal{R}_{\theta }$, as shown in Fig.\ref{fig:sgraformer}, which consists of a Scalable Graph Convolution Transformer (SGCT) and a Pose Refinement Module (PRM). 

\noindent{\bf{Scalable Graph Convolution Transformer (SGCT).}} Previous work\cite{zou2021modulated,kang2023double} has focused on learning the spatial relationships of 2D, but ignored the uncertainty from 2D to 3D. We propose to learn the latent features of 3D pose by combining the uncertain factors to obtain a more accurate 3D representation. To this end, we introduce the graph convolution transformer\cite{zhao2022graformer} to learn the latent features of 3D pose representation. Meanwhile, we combine the certain initial 3D pose with the uncertain noisy pose as input, and use 2D pose and timestep as conditions to scale and shift\cite{peebles2022scalable} the graph convolution transformer.


\noindent{\bf{Pose Refinement Module (PRM).}} Through the SGCT module, we obtain an intermediate pose containing uncertain factors. In order to better combine the certain initial pose with it, PRM obtains two weight vectors $\delta  $ and $(1-\delta )$ by training a multi-layer perceptron, which are used to balance the certain initial pose and the uncertain intermediate pose, and finally obtain the refined 3D pose.

\subsection{Multi-Hypothesis Generation and Aggregation}
\label{ssec:mhg}


To extend the multi-hypothesis prediction of our framework, we generate multiple hypotheses by combining different noises. As shown in Fig.\ref{fig:sgraformer}, it combines $H$ different noises to generate $H$ hypotheses. However, directly using multiple noises to generate hypotheses, the distribution of the results is similar. Therefore, we use $K$ times iteration of the diffusion model to enhance the differences between hypotheses. For the practical application of multi-hypothesis, we also use the aggregation method\cite{shan2023diffusion}, which maps the 3D pose hypotheses to the 2D space to obtain the closest joint, and finally obtains the best 3D pose.

\section{Experiments}
\label{sec:exp}

\subsection{Datasets and Evaluation Metrics}
\label{ssec:data}
\noindent{\bf{Human3.6M (H3.6M)}}\cite{ionescu2013human3} is an indoor dataset for 3D human pose estimation. According to the standard protocol, the model is trained on 5 subjects (S1, S5, S6, S7, and S8) and tested on 2 subjects (S9 and S11). Following \cite{ci2023gfpose,kang2023double}, Mean Per Joint Position Error (MPJPE) and Procrustes MPJPE (P-MPJPE) are used as metrics.

\noindent{\bf{MPI-INF-3DHP (3DHP)}}\cite{mehta2017monocular} has more complex cases, including studio with green screen (GS), studio without green screen (noGS), and outdoor (Outdoor). We report the Percentage of Correctly estimated Keypoints (PCK) with a threshold of 150 mm.

\subsection{Implementation Details}
\label{ssec:details}


We implement our method using Pytorch\cite{paszke2019pytorch}. The initial predictor in this paper uses DC-GCT\cite{kang2023double} to obtain the initial 3D pose. The model uses 2D joints detected by CPN\cite{chen2018cascaded} on H3.6M. The model is trained for 30 epochs with a batch size of 512. The initial learning rate is 0.0005, and a decay factor of 0.95 is applied after each epoch, with a decay rate of 0.5 every 5 epochs. We set the maximum diffusion timestep to 1000 and the sampling timestep to 200. For the cosine noise scheduler, we set the offset to 0.008.

\begin{table}[thb]
	\small
\centering
\caption{Multi-hypothesis results on the H3.6M dataset. $\textit{H}$ denotes the number of hypotheses. The top two results are bold and underlined, respectively.}
\label{table:multi}
\resizebox{.9\linewidth}{!}{
\begin{tabular}{lccc}
\toprule
Method &$\textit{H}$ &MPJPE$\downarrow$  &P-MPJPE$\downarrow$  \\
\midrule
Li \textit{et al.} \cite{li2020weakly}BMVC'20  &10 &73.9 &44.3  \\
Li \textit{et al.} \cite{li2019generating} CVPR'19  &5 &52.7 &42.6  \\
Ci \textit{et al.} \cite{ci2023gfpose} CVPR'23  &10 &\underline{45.1} &-  \\
\rowcolor{mygray}Our &10&\bf{41.8} &\bf{33.7}\\
\midrule
Sharma \textit{et al.} \cite{sharma2019monocular} ICCV'19  &200 &46.8 &37.3  \\
Oikarinen \textit{et al.} \cite{oikarinen2021graphmdn} IJCNN'21  &200 &46.2 &36.3  \\
Wehrbein \textit{et al.} \cite{wehrbein2021probabilistic} ICCV'21  &200 &44.3 &32.4 \\
Ci \textit{et al.} \cite{ci2023gfpose} CVPR'23  &200 &\underline{35.6} &\underline{30.5}  \\
\rowcolor{mygray}Our &200&\bf{35.5} &\bf{28.6}\\

\bottomrule

\end{tabular}
}
\end{table}

\begin{table}[thb]
	\Huge
\caption{Single-hypothesis results on the H3.6M dataset. $\ddagger$ indicates the deterministic model. * indicates the probabilistic model. DT and GT indicate the detected 2D poses and Ground Truth 2D poses as input, respectively. The top two results are bold and underlined, respectively.}
\label{table:single}
\centering
\resizebox{\linewidth}{!}{
\begin{tabular}{cccc}
\toprule
Method&MPJPE(DT)$\downarrow$ &P-MPJPE(DT)$\downarrow$&MPJPE(GT)$\downarrow$  \\
\midrule
Pavllo \textit{et al.} \cite{pavllo20193d}$\ddagger$  &51.8 &40.0 &37.2  \\
Zou \textit{et al.} \cite{zou2021modulated}$\ddagger$  &49.4 &39.1 &37.4  \\
Cai \textit{et al.} \cite{cai2023htnet}$\ddagger$  &48.9 &39.0 &34.0 \\
Kang \textit{et al.} \cite{kang2023double}$\ddagger$  &\underline{48.4} &\underline{38.2} &\underline{32.4}  \\
\hdashline

Wehrbein \textit{et al.} \cite{wehrbein2021probabilistic}*  &61.8 &43.8 &-  \\
Oikarinen \textit{et al.} \cite{oikarinen2021graphmdn}*  &59.2 &45.6 &-  \\
Ci \textit{et al.} \cite{ci2023gfpose}*  &51.9 &- &-  \\
\rowcolor{mygray}Our* &\bf{47.9} &\bf{38.1}&\bf{30.5} \\

\bottomrule

\end{tabular}
}
\end{table}



\subsection{Comparison with State-of-the-Art}
\label{ssec:sota}

\noindent{\bf{Comparison on H3.6M.}} Our model achieves SOTA results in both multi-hypothesis and single-hypothesis predictions. As shown in Table \ref{table:multi}, our model achieves MPJPE of 41.8 and 35.5 at $\textit{H}=10$ and $\textit{H}=200$, respectively. Especially at $\textit{H}=10$, our model reduces the MPJPE by 3.3mm compared to the SOTA method\cite{ci2023gfpose}, which is an improvement of {\bf{7.3\%}}. As shown in Table \ref{table:single}, our model achieves SOTA results on both DT and GT. Since our method is to refine the certain initial pose, our results are far better than the previous probabilistic models. Our MPJPE on DT is 47.9mm, which is 4.0mm lower than the MPJPE of the SOTA method\cite{ci2023gfpose}, which is an improvement of {\bf{7.7\%}}. More importantly, our results are even better than the deterministic model, which indicates that our model can effectively learn the latent features of the initial 3D pose, thus improving the performance. Our MPJPE on GT is 30.5mm, which is 1.9mm lower than the MPJPE of the SOTA method\cite{kang2023double} and an improvement of {\bf{5.9\%}}.

\begin{table}[!t]
\centering
\caption{Qualitative results on the 3DHP dataset. The top two results are bold and underlined, respectively.}
\label{table:3dhp}
\resizebox{\linewidth}{!}{
\begin{tabular}{ccccc}
\toprule
Method&GS$\uparrow$ &noGS$\uparrow$  &Outdoor$\uparrow$ & All PCK$\uparrow$   \\
\midrule
Li \textit{et al.} \cite{li2019generating}  &70.1 &68.2 &66.6 &67.9\\
Wehrbein \textit{et al.} \cite{wehrbein2021probabilistic}  &86.6 &82.8 &82.5 &84.3\\
Ci \textit{et al.} \cite{ci2023gfpose} &\underline{88.4} &\underline{87.1} &\underline{84.3} &\underline{86.9}\\
\rowcolor{mygray}Our &\bf{88.9}&\bf{87.9} &\bf{84.4}&\bf{87.4}\\
\bottomrule

\end{tabular}
}
\end{table}

\noindent{\bf{Comparison on 3DHP.}} We evaluate our DRPose framework on the 3DHP dataset to assess the generalization ability. We train our model on the H3.6M training dataset and test on the 3DHP test dataset. As shown in Table \ref{table:3dhp}, our method outperforms the previous methods \cite{li2019generating,wehrbein2021probabilistic,ci2023gfpose}.



\begin{table}[!t]
\small

\caption{Ablation study on the different components in DRPose framework.}
\label{table:con}
\centering
\resizebox{\linewidth}{!}{
\begin{tabular}{clcc}
\toprule
Initial Predictor& Refinement Model&Params &MPJPE$\downarrow$ \\
\midrule
\multirow{3}{*}{\makecell[c]{HTNet\cite{cai2023htnet}}} & - &3.0M  &48.9 \\
 &SGCT&4.0M  &48.6 \\
 &SGCT+PRM&4.2M  &48.3 \\
\hdashline
\multirow{3}{*}{\makecell[c]{DC-GCT\cite{kang2023double}}} & - &2.1M  &48.4 \\
 &SGCT&3.0M  &48.2 \\
 &SGCT+PRM&3.2M  &\bf{47.9} \\

\bottomrule
\end{tabular}
}
\end{table}

\subsection{Ablation Study}
\label{ssec:study}
We conducted ablation studies on H3.6M to validate the impact of each design in our framework.

\noindent{\bf{Effectiveness of each component.}} We used two 3D human pose estimation models with different accuracies, HTNet\cite{cai2023htnet} and DC-GCT\cite{kang2023double}, as initial predictors. As shown in Table \ref{table:con}, when using the same refinement model, the higher-accuracy model with better refinement results. When the initial predictor is the same, the combination of SGCT and PRM can achieve the best performance, i.e., from 48.4mm to 47.9mm when the initial predictor is DC-GCT. The proposed SGCT and PRM can effectively capture the potential 3D features and balance the certain and uncertain 3D poses.

\begin{table}[thb]
  \small
\caption{Ablation study on different configurations feasible in real-world applications. We use the single-hypothesis configuration as baseline. $\textit{H}$ denotes the number of hypotheses. $\textit{K}$ denotes the iteration times.}
\label{table:strategy}
\centering
\resizebox{.8\linewidth}{!}{
\begin{tabular}{ccccc}
\toprule
Method&Strategy&$\textit{H}$& $\textit{K}$ &MPJPE$\downarrow$ \\
\midrule
baseline&- &1 &1 &47.9 \\
\hdashline
1&\multirow{4}{*}{Average} &10 &1 &47.9{\scriptsize(-0.0)} \\
2&&10 &100 &47.7{\scriptsize(-0.2)} \\
3&&200 &1 &47.9{\scriptsize(-0.0)} \\
4&&200 &100 &47.5{\scriptsize(-0.4)} \\
\hdashline
5&\multirow{4}{*}{Aggregate\cite{shan2023diffusion}}&10 &1 &47.7{\scriptsize(-0.2)} \\
6&&10 &100 &47.5{\scriptsize(-0.4)} \\
7&&200 &1 &47.6{\scriptsize(-0.3)} \\
8&&200 &100 &{\bf{47.2}}{\scriptsize (-0.7)} \\

\bottomrule
\end{tabular}
}
\end{table}


\noindent{\bf{Different configurations in real-world applications.}} Obtaining the most suitable hypothesis from multiple hypotheses is very important in real applications. As shown in Table \ref{table:strategy}, we contrasted 9 methods that are feasible in practice. It can be clearly seen that averaging or aggregating multiple hypotheses to obtain a feasible solution can improve the result, and the method based on aggregation is better. For example, the MPJPE of method 8 is reduced by 0.7mm compared with the baseline method. In addition, it can be seen from the table that MPJPE is smaller when more hypotheses are generated (such as methods 5 and 7) or more time steps are sampled (such as methods 7 and 8) when other conditions are constant. In practice, different configurations can be balanced according to the actual situation.


\begin{figure}[!t]
\centering
\includegraphics[width=\linewidth]{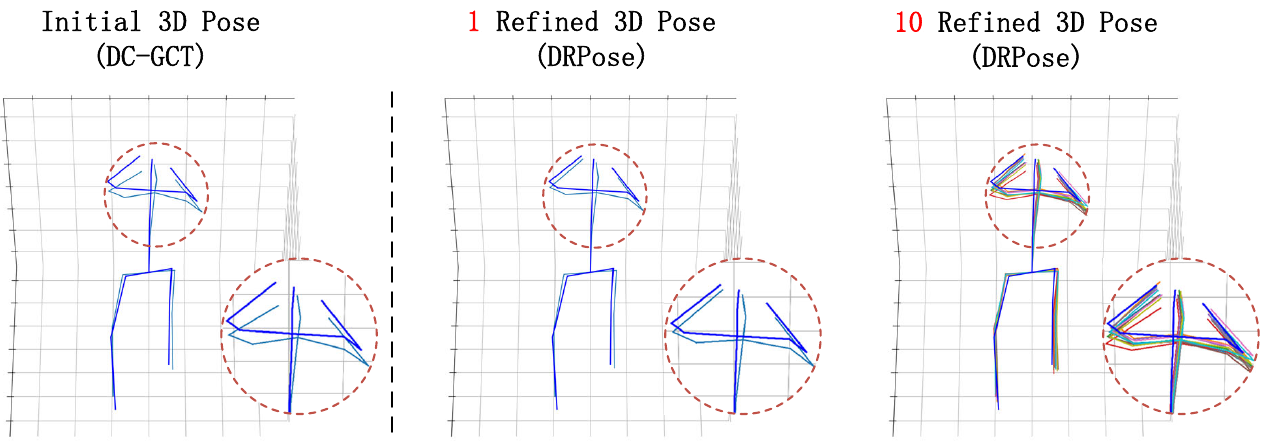}
\caption{Qualitative visual results of our method on the H3.6M test dataset. The left shows the initial 3D pose obtained by DC-GCT\cite{kang2023double}, and the right shows the single and multiple refined 3D poses obtained by our DRPose. The blue pose represents the ground truth.}
\label{fig:show}
\end{figure}

\subsection{Qualitative Results}
\label{sec:visualization}

Fig.\ref{fig:show} shows the visualization results on the H3.6M test dataset. Compared with SOTA method DC-GCT\cite{kang2023double}, our method achieves better single hypothesis prediction, which also proves that our framework can effectively refine the initial 3D pose. In addition, we obtain multiple hypotheses that are closer to the ground truth by combining multiple noise and iteratively refining. This well models the uncertainty of the 2D detector and the depth blur best.

\section{Conclusion}
\label{sec:Conclusion}

This paper presents DRPose, a diffusion-based refinement framework for improving 3D pose estimation. By combining deterministic and probabilistic predictions, leveraging SGCT for denoising and latent feature learning, and employing PRM for balancing the certain and uncertain poses, DRPose achieves enhanced accuracy in both single and multi-hypothesis scenarios. Experimental validation on benchmark datasets demonstrates its superiority, establishing DRPose as a SOTA solution for accurate 3D human pose estimation. We hope our framework can achieve better extension, such as combining a more accurate initial prediction model, or adding temporal information.




\bibliographystyle{IEEEbib}
\bibliography{refs}

\end{document}